\documentclass[sigconf]{acmart}

\AtBeginDocument{%
  }

\copyrightyear{2024}
\acmYear{2024}
\setcopyright{acmlicensed}\acmConference[CIKM '24]{Proceedings of the 33rd ACM International Conference on Information and Knowledge Management}{October 21--25, 2024}{Boise, ID, USA}
\acmBooktitle{Proceedings of the 33rd ACM International Conference on Information and Knowledge Management (CIKM '24), October 21--25, 2024, Boise, ID, USA}
\acmDOI{10.1145/3627673.3679614}
\acmISBN{979-8-4007-0436-9/24/10}

\usepackage{graphicx}
\usepackage{subcaption}
\usepackage{multirow}
\usepackage{multicol}
\usepackage{booktabs}
\usepackage{amsmath}
\usepackage{bbm}
\usepackage{tabularx}
\usepackage{enumitem}

\begin{document}

\title{Multivariate Time-Series Anomaly Detection based on Enhancing Graph Attention Networks with Topological Analysis}

\author{Zhe Liu}
\affiliation{%
  \institution{Beihang University}
  \city{Beijing}
  \country{China}
}
\email{liuzhe2339@buaa.edu.cn}

\author{Xiang Huang}
\affiliation{%
  \institution{Beihang University}
  \city{Beijing}
  \country{China}
}
\email{huang.xiang@buaa.edu.cn}

\author{Jingyun Zhang}
\affiliation{%
  \institution{Beihang University}
  \city{Beijing}
  \country{China}
}
\email{zhangjingyun@buaa.edu.cn}

\author{Zhifeng Hao}
\affiliation{%
  \institution{Shantou University}
  \city{Guangdong}
  \country{China}
}
\email{haozhifeng@stu.edu.cn}
\authornote{Corresponding author}

\author{Li Sun}
\affiliation{%
  \institution{North China Electric Power University}
  \city{Beijing}
  \country{China}
}
\email{ccesunli@ncepu.edu.cn}

\author{Hao Peng}
\affiliation{%
  \institution{Beihang University}
  \city{Beijing}
  \country{China}
}
\email{penghao@buaa.edu.cn}

\begin{abstract}
Unsupervised anomaly detection in time series is essential in industrial applications, as it significantly reduces the need for manual intervention. 
Multivariate time series pose a complex challenge due to their feature and temporal dimensions. 
Traditional methods use Graph Neural Networks (GNNs) or Transformers to analyze spatial while RNNs to model temporal dependencies. 
These methods focus narrowly on one dimension or engage in coarse-grained feature extraction, which can be inadequate for large datasets characterized by intricate relationships and dynamic changes. 
This paper introduces a novel temporal model built on an enhanced Graph Attention Network (GAT) for multivariate time series anomaly detection called TopoGDN. 
Our model analyzes both time and feature dimensions from a fine-grained perspective. 
First, we introduce a multi-scale temporal convolution module to extract detailed temporal features. 
Additionally, we present an augmented GAT to manage complex inter-feature dependencies, which incorporates graph topology into node features across multiple scales, a versatile, plug-and-play enhancement that significantly boosts the performance of GAT. 
Our experimental results confirm that our approach surpasses the baseline models on four datasets, demonstrating its potential for widespread application in fields requiring robust anomaly detection.
The code is available at \url{https://github.com/ljj-cyber/TopoGDN}. 
\end{abstract}



\maketitle
\section{Introduction}
\begin{figure}[t]
    \centering
    \includegraphics[width=\columnwidth]{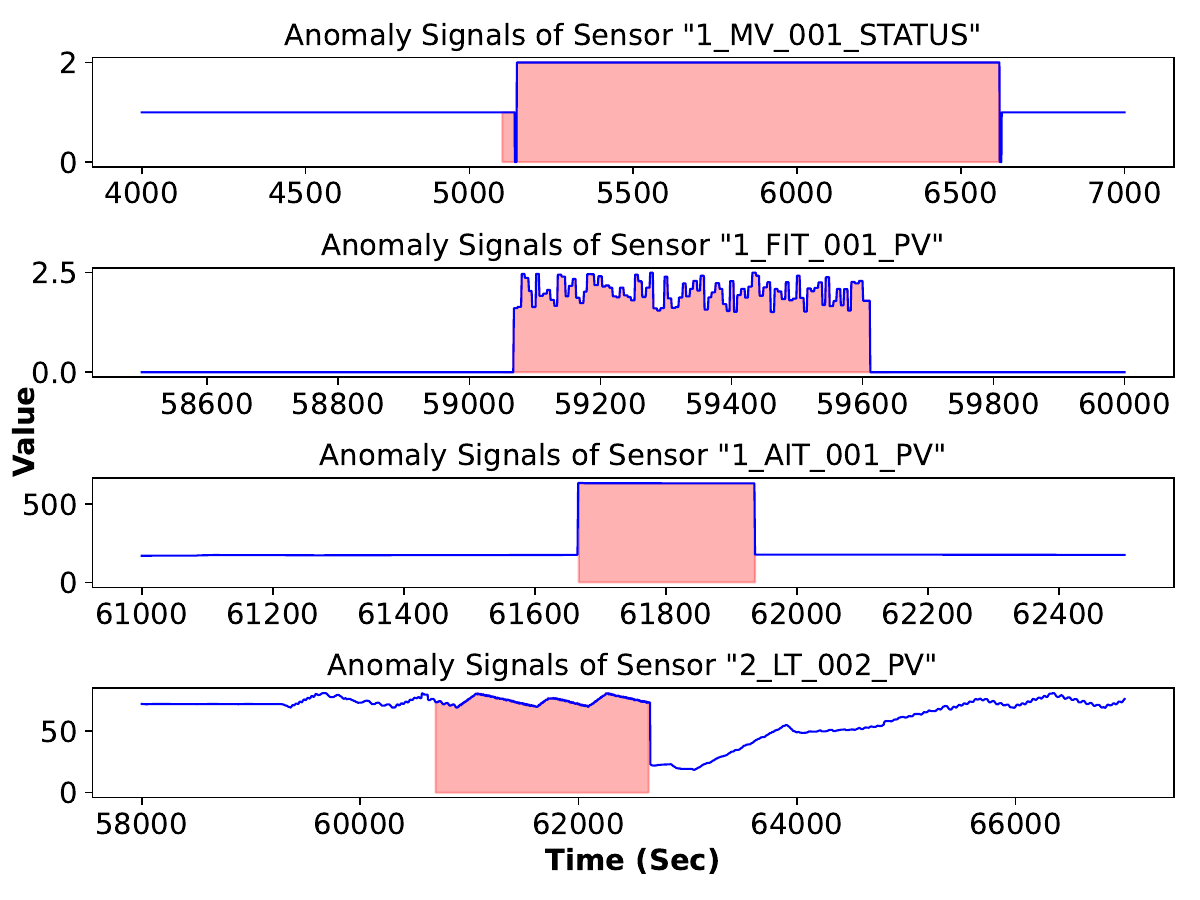}
    \caption{Anomalous Behavior in Industrial Sensor Time Series Data: 
    \textmd{The first three anomalies shown in the chart occur in specific time intervals and are characterized by sudden changes in value relative to the surrounding context. 
    The last anomaly indicates fluctuations in the water level's rise and fall rate. This anomaly exhibits a specific pattern but tends to diminish over time.}}
    \label{fig:enter-label}
\end{figure}
Time-series data, characterized by observations sampled at discrete time intervals, exhibits strong temporal correlations and often displays periodic patterns. 
These data are prevalent in various sectors, including industrial \cite{ahmed2017wadi,mathur2016swat}, medical, transportation \cite{abdulaal2021practical}, and network environments \cite{su2019robust}. 
In these applications, time-series data may appear as industrial sensor readings, network traffic, traffic flows, or medical records. 
Analyzing these time-series data includes regression and classification, typically focusing on predicting future values \cite{wu2020connecting} and detecting anomalies \cite{schmidl2022anomaly}, respectively. 
The latter is particularly crucial, as timely anomaly detection enables prompt intervention, essential in scenarios like network security breaches or traffic disruptions due to accidents or special events.
For example, Figure \ref{fig:enter-label} shows anomalous data patterns in an industrial scenario.

With the advancements in computing power, large language models have gained prominence, setting new standards across numerous tasks \cite{pangFrozenTransformersLanguage2023}. 
However, these models often struggle to effectively handle sparse features from data with distributional imbalances, such as those with long-tailed distributions characteristic of temporal anomalies. 
The increasing complexity of time series data, spurred by the proliferation of industrial Internet of Things platforms, adds complexity by introducing dependencies among features in multivariate time series \cite{feng2021time, zeng2023multivariate}. 
Additionally, the need for labeled anomaly data complicates the detection of various anomaly patterns in an unsupervised manner. 
Temporal anomalies can be divided into two categories: global anomalies, which significantly deviate from overall statistical properties, and contextual anomalies \cite{ma2021comprehensive}, which, although within a normal range, show significant deviations from expected patterns, thereby disrupting periodicity. 
Hence, effectively managing the relationship between local and global \cite{zou2024multispans} anomalies over an extended period presents a significant challenge in temporal tasks.

The initial approaches to unsupervised time series anomaly detection primarily utilized clustering \cite{scholkopf2001estimating}, density-based \cite{breunig2000lof}, and shape-based \cite{hallac2017toeplitz} methods to identify simple patterns such as global outliers. 
Although effective for basic detections, these methods suffer scalability and accuracy issues as the temporal dimension expands. 
Recent research efforts have shifted towards refining the definitions of anomaly patterns and categorizing detection methods into predictive and reconstructive approaches \cite{hanLearningSparseLatent2022}. 
Both approaches rely on comparing model outputs with the ground truth to derive anomaly scores, which are then thresholded using statistical methods to identify anomalies. 
High scores generally indicate anomalous events, necessitating further investigation. 
Prediction-based approaches excel in utilizing time dependencies and adapting to time series dynamics, yet they are susceptible to new patterns and heavily reliant on historical data. 
On the other hand, reconstruction-based methods offer a more holistic representation of data and robust non-linear modeling capabilities \cite{hanLearningSparseLatent2022}. 
However, higher computational demands challenge them, and the representations in the latent space may substantially diverge from the true meanings of the actual data.
Compared to univariate time series anomaly detection, the primary challenge in multivariate detection is effectively modeling the complex dependencies among multiple sequences. 
Trying to model these dependencies in some instances can adversely impact predictive accuracy, particularly when the statistical characteristics of different features diverge significantly. 
Merging such features might inadvertently introduce noise \cite{chen2022cross}. 
However, failure to adequately manage these inter-feature dependencies can lead the model to settle into suboptimal solutions. 
To solve the above problems, GAT \cite{velivckovic2017graph} is well-suited for addressing dependency extraction challenges, which can leverage the structural features of constructed graphs, using neighborhood information instead of global data to filter out irrelevant dependencies more effectively.

Inspired by GDN \cite{deng2021graph}, we propose an enhanced GAT based on topological analysis called TopoGDN, which models dependencies across time and features simultaneously at multiple scales. 
Our prediction-based approach utilizes GAT and consists of four main components.
The first is a multi-scale time series convolution module that extracts temporal features at various scales before inputting them into the GNN. 
The second is a graph structure learning module, which constructs graph structures using the similarities between sequences and continuously updates these structures as the time window progresses. 
The third is the topological feature attention module, which extracts higher-order topological features from graph structures and integrates them with the temporal features. 
Lastly, the anomaly scoring module calculates deviation scores by comparing current time steps with normal baselines, using predefined statistical thresholds to identify anomalies. The contributions of our study are summarized as follows:
\begin{itemize}[left=0pt]
\item We propose TopoGDN, an anomaly detection framework that extracts temporal and inter-feature dependencies separately across multiple scales.
\item We design a multi-scale time series convolution module to address the conflict between local and global temporal features within an extended time window.
\item We propose an enhanced GAT that models higher-order topological features as persistent homology groups following varying degrees of graph filtering. 
This approach effectively improves the accuracy of modeling inter-feature dependencies.
\item Experimental validation has demonstrated that our method outperforms baseline models on four benchmark datasets, showcasing its broad application potential.
\end{itemize}
\section{Related Work}
This section briefly reviews earlier anomaly detection methods for multivariate time series data. 
Then, we introduce two techniques in our approach: graph neural network and topological analysis.
\subsection{Multivariate Time Series Anomaly Detection}
Multivariate time series anomaly detection involves classifying anomalies across feature and temporal dimensions.
Existing anomaly detection methods are classified into two categories based on how the data is processed in the model: prediction and reconstruction-based.
The prediction-based approach generates future data by learning from historical data and comparing the generated and actual data to determine anomalies. 
GDN \cite{deng2021graph} first utilizes similarity to model graph structures, employing graph-based techniques to map dependencies among features. 
However, this approach suffers from inadequate utilization of temporal dependencies.
Anomaly Transformer \cite{xuAnomalyTransformerTime2021} leverages the Transformer to model dependencies between features and employs a variable Gaussian kernel to model temporal dependencies. 
However, the Transformer struggles to cope with anomaly-sparse scenarios.
Our method also adopts a prediction-based approach, separately modeling both dependencies.

Reconstruction-based methods, on the other hand, use generative models such as Variational Auto-Encoder (VAE) \cite{an2015variational} and Generative Adversarial Network (GAN) \cite{goodfellow2020generative} to model data representations.
Lin et al. \cite{lin2020anomaly} leverages the VAE to create robust local features in short windows and utilizes the LSTM to estimate long-term correlations.
MAD-GAN \cite{li2019mad} employs the GAN's generator to learn implicit features of time series, while the discriminator identifies genuine and fake sequences and detects anomalies.
The limitations of these generative models include the instability of GAN training, heightened sensitivity to hyperparameters, increased risks of non-convergence in training, and blurred generated data due to the smooth latent space of VAE.

\subsection{Graph Neural Network}
In previous studies, Graph Neural Networks (GNNs) are often employed to address graph structure-related tasks, such as node classification \cite{xiao2022graph} and edge prediction \cite{zhanglink2018}.
Among these, Graph Convolutional Networks (GCNs) \cite{defferrard2016convolutional}, which leverage spectral domain transformations, are at the forefront. 
GCN integrates convolution operations from traditional Convolutional Neural Networks (CNNs) with the Laplacian graph matrix that facilitates message diffusion. 
Subsequently, as the understanding of graph structures has evolved beyond mere predefined knowledge, graph structure learning emerges to model similarities between variables. 
This has led to an expansion in the application areas of graph neural networks, such as multivariate time series analysis. 
GAT \cite{velivckovic2017graph} introduces an attention mechanism to define relationships between nodes, assigning learnable coefficients to each edge. 
Unlike GCNs, GATs can dynamically adjust these attention coefficients to accommodate changes in the graph structure. Research has explored the connections between GATs and Transformers \cite{ying2021transformers}, demonstrating that the attention coefficient matrix computed by GAT tends to be sparser than those in traditional attention maps, an attribute that helps explain the sparsity observed in anomalous data.

\subsection{Topological Analysis}
Topological data analysis comprises methods that encapsulate structural information about topological features, including connectivity components, rings, cavities, and more \cite{carriere2020perslay}. 
Using a persistence diagram, the statistical properties of the above topological features can be mapped onto a two-dimensional plane and finally vectorized. 
Persistence Homology (PH) is a topological analysis method demonstrating that graph filtering functions, which remain invariant under graph isomorphism, can assimilate information about edges and nodes into a near-simplicial complex \cite{horn2021topological}. 
Topological features serve as higher-order structural elements to address the over-smoothing issue in GNN \cite{zhou2022hypergraph}. 
Pho-GAD \cite{yuan2024phogad} enhances the distinction between normal and abnormal node behaviors by optimizing edge attributes using persistent homology. 
Chen et al. \cite{chen2021topological} leverage the persistent homology of a node's immediate neighborhood to rewire the original graph, incorporating the resultant topological summaries as auxiliary data into local algorithms. 
Meanwhile, Zhou et al. \cite{zhou2023overcoming} employ topological analysis to refine the understanding of complex human joint dynamics using the graph's adjacency matrix.
\section{Methodology}

\begin{figure*}[ht]
    \centering  \includegraphics[width=\textwidth]{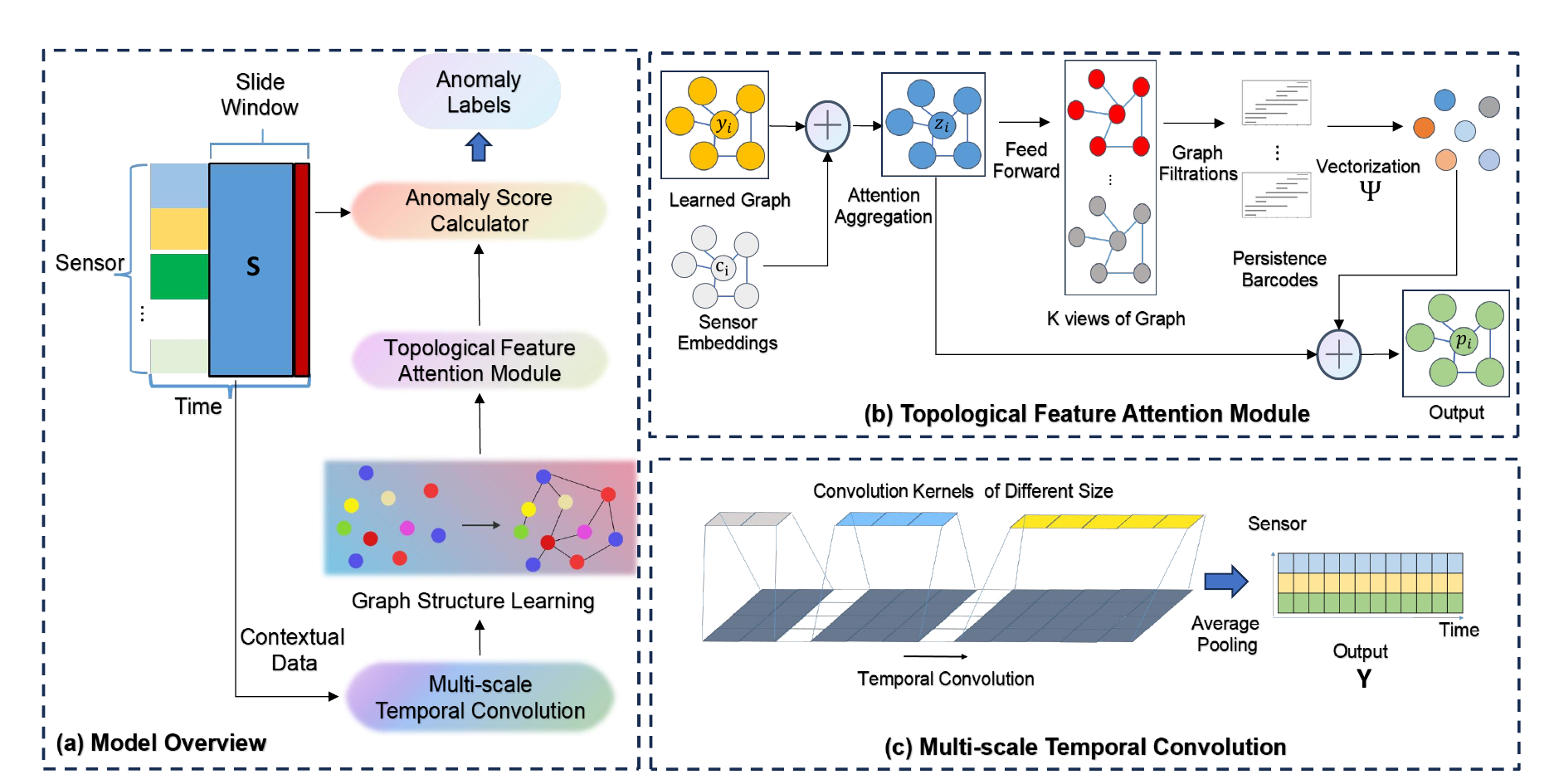}
   \caption{\textit{TopoGDN} model structure and local module diagrams. \textmd{
   As shown in (a) Model Overview, we first employ a sliding window approach to select contextual data from sensor inputs in the initial stage. 
   This data is processed using the (c) Multi-scale Temporal Convolution module, which applies convolutional kernels of varying sizes to capture temporal features. 
   These extracted features are then passed to the Graph Structure Learning module, transforming them into a contextual graph.
   Next, the (b) Topological Feature Attention Module utilizes sensor embeddings and the learned graph to perform attention aggregation. 
   The module then conducts graph filtrations, creating multiple views of the graph to enhance feature representation. 
   The resulting outputs are vectorized and converted into persistence barcodes, which capture essential topological features.
   Finally, the anomaly score is calculated by comparing the predicted outputs with the ground truths.
   }
}
    \label{fig:image1}
\end{figure*}
In this section, we provide a detailed overview of the proposed TopoGDN framework. 
We begin by outlining the problem formulation for multivariate time series anomaly detection, followed by describing our data preprocessing methodology. 
Next, we introduce and discuss the modules integrated into our approach. 
Specifically, our method comprises the following four modules:
\begin{enumerate}[left=0pt]
    \item \textbf{Multi-scale Temporal Convolution Module}: This module extracts temporal features at multiple scales and incorporates residual connections to enhance the learning from data within the time window.
    \item \textbf{Graph Structure Learning Module}: It constructs dynamic graph structures based on the distributional similarity among variables, continuously adapting as the window progresses.
    \item \textbf{Topological Feature Attention Module}: This module aggregates embeddings from individual nodes, their neighboring nodes, and higher-order topological features, fusing inter-feature dependencies at multiple scales.
    \item \textbf{Anomaly Score Calculator}: This module computes an anomaly score for the predicted time step, aiding in identifying potential outliers.
\end{enumerate}

\subsection{Problem Formulation}
Given a dataset $\mathbf{X} \in \mathbb{R}^{N \times T}$, where $N$ represents the number of features and $T$ denotes the number of time steps. 
We distinguish between a training set $X_{\text{train}}$, which consists entirely of regular data, and a test set $X_{\text{test}}$ that includes both normal and abnormal data.
The model aims to predict anomaly scores $s_t$ for each time step $t$ in the test set. 
These scores are then converted into labels $\mathbf{Y}_{\text{test}}$ based on the proportion of anomalies present in the dataset. 

\subsection{Data Preprocessing}
The values across different columns vary significantly in magnitude in the original dataset. 
For example, in the MSL dataset, most data are binary, with values being either 0 or 1. 
To mitigate the impact of different scales on the final prediction results, we employ min-max normalization to map all values to the interval [0, 1]. 
The formula is given by:
\begin{equation}
X_t^{i} = \frac{X_t^{i} - \min(X^{i})}{\max(X^{i}) - \min(X^{i}) + \epsilon}, 
\end{equation}
where $X^{i}$ is the feature vector with all time steps, and $\epsilon$ is a small vector added to avoid division by zero in cases where $X^{i}$ contains identical values.
Our model employs a prediction-based anomaly detection approach using a sliding time window to divide the dataset into training and test sets. 
We define the window as a stride length of \( s \) and a size of \( w \). 
This configuration allows the window to be shifted by \( s \) units at each step, providing a mechanism for sequential data processing across the dataset.
The \textit{Slide Window} operation is defined such that for any given time step $t$, the window $S = [X_{t-w}, X_{t-w+1}, \ldots, X_{t-1}]$ is constructed, where $S \in \mathbb{R}^{N \times w}$. 
The window predicts the next data point $\hat{X}_t$. 
The model learns from the data of the previous $w$ steps to predict the data at the next time step $t$, denoted as $\hat{X}_t$.
By comparing $\hat{X}_t$ with $X_t$, it determines whether the time step $t$ is anomalous. 
The sliding window continuously moves forward. 
If the final data segment is shorter than the step length, we use replication padding with the last step's data.
The model is solely trained on its predictive capabilities in the training set, while scores are predicted only on the test set.

\subsection{Multi-scale Time Convolution}
To address the conflict between local and global features over a long period, we design a multi-scale temporal convolution module to capture fine-grained temporal dependencies.
This module features one-dimensional convolution kernels of various sizes, which capture temporal features at different scales in the time series.
These convolutions may be dilated to increase their receptive field without increasing the kernel size \cite{zou2024multispans}. 
Given a time series \(S^{i} \in \mathbb{R}^{w}\), we employ \(P\) different sizes of convolution kernels. 
Each kernel size \(w_p\) contributes to extracting features at a particular scale. 
The convolution operation for the \(p\)-th kernel is defined as follows:
\begin{equation}
y^{(p)}[t] = \sum_{j=0}^{w_p-1} S^{i}[t+j] \cdot f_p[j],
\end{equation}
where \(y^{(p)}[t]\) represents the output at time \(t\) using the \(p\)-th convolution kernel, \(f_p[i]\) is the weight of the \(p\)-th kernel at position \(i\), and \(w_p\) is the width of the \(p\)-th kernel.
After applying the convolution kernels, the module performs a pooling operation across the outputs from all kernels to synthesize and enhance the representation of temporal dynamics. 
This step typically uses average pooling to merge the diverse features extracted by each kernel into a unified output. 
The final output \(Y^i\) from the pooling layer is then used in subsequent processing:
\begin{equation}
Y^i[t] = \mathbf{pool}(y^{(1)}[t], y^{(2)}[t], \ldots, y^{(P)}[t]),
\end{equation}
where \(\mathbf{pool}(\cdot)\) is the pooling function employed to combine the outputs from each convolution scale into a single vector, effectively capturing temporal features across different scales and maintaining the same dimensionality as the input \(S^i\).
The convolution kernels used in this implementation include sizes \(1 \times 2 \), \(1 \times 3 \), \(1 \times 5\), and \(1 \times 7\), designed to accommodate a broad range of temporal resolutions.

\subsection{Graph Structure Learning}
Drawing on insights from GDN, it is crucial to model a graph structure through similarity computations when dealing with multivariate data that lacks inherent structure.
In the proposed graph model, we represent each sensor within the dataset as a node.
We construct edges between nodes based on the similarity of the data corresponding to the features they represent.
Specifically, the probability of an edge between two nodes increases proportionally with the similarity observed in their associated data.

Consider a learned graph denoted as \( G = (V, E) \), where \( V \) represents the set of nodes with their embeddings given by \( Y \in \mathbb{R}^{N \times w} \), capturing feature representations for each node. 
Using an adjacency matrix, we depict the relationships between nodes, which encodes their connectivity and interaction details.
We assess the similarity between the data of the two entities by employing a neural network to refine the entity embeddings.
Initially, each entity is assigned a random embedding \( c_i \in \mathbb{R}^d \). 
This embedding vector serves dual purposes: firstly, to represent the data features of the entities within a new embedding space, facilitating the evaluation of inter-entity similarities. 
Secondly, it introduces new data features during subsequent graph attention aggregation. 
The embeddings \( c_i \) are trained on the training set, and once this training phase is complete, the embeddings are fixed—meaning the graph structure remains unchanged for the test set.
We employ the dot product to calculate similarity, utilizing pairwise entity comparisons. 
Thus, the adjacent matrixes are calculated as:
\begin{equation}
    A_{ij} = \mathbbm{1}\{j \in \text{Top-K}(\{e_{ki} \})\},
\end{equation} 
\begin{equation}
    e_{ij} = \frac{\mathbf{c}_i^T \cdot \mathbf{c}_j}{\|\mathbf{c}_i\| \cdot \|\mathbf{c}_j\|} \quad.
\end{equation} 
Notably, we use condition function $\mathbbm{1}\{\cdot\}$ to indicate that the element ${A_{ij}}$ of the adjacency matrix ${A}$ is 1 if vertex $j$ is among the top K largest values of ${e_{ki}}$ related to vertex ${i}$; otherwise, it is 0.
Given that our approach integrates an attention network, the neighbor matrix is binary, composed solely of 0s and 1s. 
In our graph attention update mechanism, there is no need to incorporate the node's features repeatedly. Consequently, the computed edge information excludes self-loops from the nodes.

\subsection{Topological Feature Attention Module}
As shown in Figure \ref{fig:image1} (b), the Topological Feature Aggregation module consists of two key components: the graph attention mechanism and topological graph pooling. 
Initially, node aggregation occurs through the graph attention mechanism, facilitating the integration of local and global information across multiple layers. 
Subsequently, the module extracts high-order topological structures from the graph, converting them into a Persistence barcode. 
The structural information is then globally vectorized and fused, resulting in the final feature graph. 

\subsubsection{Graph Attention Mechanism}
We have adopted the Graph Attention Mechanism from GATs, which aggregates information from neighboring nodes by computing attention coefficients for pairs of adjacent nodes. 
We divide the output of the Multi-Scale Temporal Convolution module $Y$ such that the feature vector corresponding to each sensor becomes node embedding $Y^i \in \mathbb{R}^w$.
Before calculating the attention coefficients $\alpha_{ij}$, the following operations are performed:
\begin{equation}
    h_i = W \cdot Y^i,
\end{equation}
\begin{equation}
    g_i = c_i \oplus h_i,
\end{equation}
\begin{equation}
    s_{ij} = \mathbf{FeedForward}([g_i||g_j]).
\end{equation}
Notably, each node's embedding is transformed to obtain the hidden embedding $h_i \in \mathbb{R}^{d}$.
Here, \(W\) is a shared parameter matrix incorporating node features. 
The intermediate vectors \( g_i \) and \( g_j \), resulting from the fused sensor embeddings, are combined and then transformed into an actual number using a feed-forward neural network.
This process serves as a typical feature enhancement technique. 
This attention mechanism bears a resemblance to the self-attention mechanism found in Transformers \cite{vaswani2017attention}. 
Additionally, optimizing it using a multi-head mechanism allows for more nuanced and practical learning.
The computation of the attention coefficients involves two main steps:
\begin{equation}
    \alpha_{ij} = \frac{\exp(\mathbf{LeakyReLU}(s_{ij}))}{\sum_{k \in \mathcal{N}(i)} \exp(\mathbf{LeakyReLU}(s_{ik}))},
\end{equation}
\begin{equation}
    z_i^{(t)} = \sigma(\sum_{j\in{N_i}}\alpha_{ij}h_j).
\end{equation}
We scale these coefficients appropriately by applying the LeakyReLU activation function and Softmax normalization.
The final step involves message aggregation, where the hidden layer feature of a node, denoted as $h_i$, is attentively weighted and summed with the features of its neighboring nodes to derive the final node feature $z_i$.
Notably, \( N(i) \) represents the neighbors of node \( i \), and \( \sigma(\cdot) \) refers to the Sigmoid activation function.

\subsubsection{High-order Topological Graph Pooling}
In algebraic topology, persistence homology tools aim to identify connected components (0-dimensional homology) and cycles (1-dimensional homology) within graph homology. 
This method has proven effective in optimizing graph representation learning \cite{hofer2019learning,hofer2020graph}.

We apply \textbf{High-order Topological Graph Pooling} to the node embeddings \( \mathbf{Z} \in \mathbb{R}^{N \times d} \) and the graph structure information, which the graph attention mechanism has refined. 
In this process, \( N \) represents the number of vertices, and \( d \) the dimensionality of each vertex's hidden embedding.
This pooling operation aggregates and abstracts the enhanced feature representations, succinctly capturing the topological characteristics of the graph.
Initially, the nodes undergo dimensionality increase using a multi-layer perceptron.
After the dimension transformation, we describe the graph from $k$ different perspectives, resulting in \( k \) transformed graphs \( K = \{K_1, K_2, \ldots, K_k\} \in \mathbb{R}^{k \times N \times d} \). 
This process allows for the extraction of richer topological features at multiple scales.
We employ a set of \( k \) filtering functions to perform graph filtering from \( k \) distinct viewpoints. 
Specifically, we apply graph filtering to each transformed graph \( K_i \) (for \( i = 1, \ldots, k \)), which processes the information of nodes and edges for each viewpoint.
We filter the graph according to this form:
\begin{equation}
    \emptyset = K_i^{(0)} \subseteq K_i^{(1)} \subseteq \cdots \subseteq K_i^{(n)} = K_i,
\end{equation}
\begin{equation}
    K_i^{(j)} = (V_i^{(j)}, E_i^{(j)}).
\end{equation}
The subscript \(i\) in the formula represents the graph of the \(i\)-th view; the superscript \((j)\) represents the degree of filtration, where the closer \(j\) is to 0, the more extensive the filtration. 
The \(z_v\) represents the embedding of the node \(v\), and \(f_i(\cdot)\) represents the pooling function done on the node embedding in the graph of the \(i\)-th viewpoint. 
Our designed graph filtering method uses threshold filtering, where nodes above the threshold \(\alpha_i^{(j)}\) and some of their connecting edge information are removed in this form:
\begin{equation}
    V_i^{(j)} = \left\{ v \in V_i^{(j+1)} \mid f_i(z_{v}) \leq a_i^{(j)} \right\},
\end{equation}
\begin{equation}
    E_i^{(j)} = \left\{ (u, w) \in E_i^{(j+1)} \mid \max \{f_i(z_{u}), f_i(z_{w})\} \leq a_i^{(j)} \right\}.
\end{equation}
Since Vietoris-Rips complexes better represent higher-order topological features, we apply Vietoris-Rips complexes to construct simple complexes from graphs as higher-order topological features of the graphs, and only 0- and 1-dimensional complexes, i.e., connected components and voids, are considered in our implementation \cite{aktas2019persistence}. 
A tuple represents the persistence of a topological feature in graph filtering. 
If a simplicial complex appears in \(K_i\) and disappears in \(K_j\) after filtering, we can represent it by \((i,j)\), which corresponds to the value of \(j - i\). 
This collection of two-dimensional tuples is the Persistence Diagram:  
\begin{equation}
    D^{(l)} = \{(i, j) \mid \text{for each } S \text{ in Simplex}(G')\},
\end{equation}
where \(i\) is the birth time of the simplicial complex \(S\) and \(j\) is the death time of \(S\).
\( l = 0 \) denotes the 0-dimensional homology, known as connected components, while \( l = 1 \) indicates the 1-dimensional homology for cycles.
The Persistence Barcode is an intuitive visual representation of the Persistence Diagram showing topological features' appearance and disappearance. 
Each bar represents a topological feature, such as a simplicial complex like a connected component, a hole, or a void.
The start and end spans of the bar represent the time frame in which the feature exists. 
The horizontal axis corresponds to the degree of filtration. 
In contrast, the vertical axis has no real mathematical significance but is used to arrange the different bars, each representing a specific topological feature.
Finally, we use the embedding function $\Psi$ to transform the Persistence Barcode into Topological Vectors.
This process can be formulated as follows: 
\begin{equation}
    \Psi^{(l)} : \{D_1^{(l)}, \ldots, D_k^{(l)}\} \rightarrow \mathbb{R}^{N \times d},
\end{equation}
where k is the number of filtration functions and the number of graphs' views. 
In the implementation, we use four different embedding functions to deal with multi-view graphs: (i) the \textit{Triangle Point Transformation}, (ii) \textit{Gaussian Point Transformation}, (iii) \textit{Line Point Transformation}, (iv) \textit{Rational Hat Transformation}.
Please refer to \cite{hofer2019learning} and \cite{carriere2020perslay} for details.  

\subsection{Anomaly Score Calculator}
Suppose the node embeddings of the graph after the Topological Feature Attention Module are denoted as \( p^{(t)} \in \mathbb{R}^{N \times d} \). 
After the dimensional transformation, the predicted time step data \(\hat{X_t} \in \mathbb{R}^N\) is obtained. 
This process is  formulated as: 
\begin{equation}
    \hat{X_t} = \mathbf{FeedForward}(p^{(t)}),
\end{equation}
where a feed-forward neural network is employed to achieve this transformation. 
This dimensional transformation is equivalent to performing a pooled dimensionality reduction.
On the training set, we update the weights of the sensor embedding and other neural networks based on the average mean square error of \( X_t\) and \( \hat{X_t} \) at each step as the backward loss without computing anomaly scores. 
On the test set, we similarly compare the prediction result \( \hat{X_t} \) with the ground truth \( X_t \), compute its L1-Loss, and normalize the difference. 
After normalization, we use the maximum value in the sensor dimension as the anomaly score, calculated as: 
\begin{equation}
    \text{Anomaly Score} = \max_{i=1,...,N} \left( \text{Normalize}\left( \left| \hat{X}_t^i - X_t^i \right| \right) \right),
\end{equation}
where $X_t^i$ is the value taken by the ith sensor at time t.
Finally, if the anomaly score exceeds the preset threshold, the time step \( t \) is marked as an anomaly. 
In our implementation, we adopted the method from GDN, setting the threshold as the maximum anomaly score observed on the validation set. 
The ratio of the training set to the validation set closely matches the proportion of anomalies in the dataset.
\section{Experiments}
This section elaborates on the experiment and the details of the model implementation.
\subsection{Datasets}
We select four of the most commonly used datasets for time series anomaly detection, which are:
\textbf{Mars Science Laboratory rover (MSL)} \cite{hundman2018detecting}, 
\textbf{Secure Water Treatment (SWaT)} \cite{mathur2016swat},
\textbf{Server Machine Dataset (SMD)} \cite{su2019robust},
\textbf{Water Distribution (WADI)} \cite{ahmed2017wadi}
Among them, the WADI dataset has the largest volume of data to detect and is consequently the most challenging. 
The MSL dataset has the smallest feature dimension, making detection less complicated. 
Table \ref{tab:Dataset} details our four benchmark datasets, including their sizes and Anomaly Rates (\%).

\begin{table}[h]
\caption{Statistics of datasets.}
\begin{tabular*}{\linewidth}{l|cccc}
\toprule
Dataset & Dimension & \#Training & \#Testing & Anomaly Rate \\
\midrule
MSL  & 25 & 58,317  & 73,729  & 10.50 \\
SWaT & 51 & 47,515  & 44,986  & 11.97 \\
SMD  & 38 & 23,688  & 23,688  & 4.20  \\
WADI & 127 & 118,795 & 17,275  & 5.99  \\
\bottomrule
\end{tabular*}
\label{tab:Dataset}
\end{table}

\subsection{Baselines}
To assess the performance of TopoGDN, we compare it to seven methods that belong to three categories: traditional machine learning approaches, reconstruction-based methods, and prediction-based methods.
\begin{itemize}[left=0pt]
    \item \textbf{PCA} \cite{mackiewicz1993principal}: PCA is a traditional machine learning technique that finds a dimensionality reduction mapping to extract features from data, using reconstruction error as the Anomaly Score.
    \item \textbf{LSTM-VAE} \cite{park2018multimodal}: LSTM-VAE utilizes the VAE for data reconstruction combined with the LSTM to extract temporal features.
    \item \textbf{MAD-GAN} \cite{li2019mad}: MAD-GAN employs the GAN for data reconstruction, using the generator to learn implicit features of time series and the discriminator to directly identify genuine and fake sequences, enabling it to detect anomalies.
    \item \textbf{OmniAnomaly} \cite{su2019robust}: OmniAnomaly utilizes stochastic variables linkage and regularization flow techniques from probabilistic graphical models to capture the typical patterns of data, determining anomalies by reconstructing probability distributions.
    \item \textbf{GDN} \cite{deng2021graph}: GDN uses graph structure learning to model dependencies among features and employs the GAT to extract temporal features, representing a typical prediction-based method for anomaly detection.
    \item \textbf{TranAD} \cite{tuli2022tranad}: TranAD utilizes a conventional Transformer as a sequence encoder, designing an adversarial-based training process to amplify reconstruction errors and employing model-independent meta-learning to accelerate inference.
    \item \textbf{ImDiffusion} \cite{chen2023imdiffusion}: ImDiffusion utilizes an interpolation-based approach for data reconstruction, leveraging information about neighboring values to extract complex timing-dependent information using Diffusion \cite{ho2020denoising} model.
\end{itemize}
Among them, GDN, TranAD, and ImDiffusion are advanced methods developed in the last three years, and they serve as benchmarks for our comparison.

\subsection{Experimental Setup}
We conduct all experiments on an NVIDIA GeForce 3090 with 24GB of memory. 
The programming language and framework used for the implementation are Python 3.7, PyTorch 1.13.1, CUDA 11.7, and PyTorch-Geometric 1.7.1.
Default generic hyperparameters are presented in Table \ref{table: Hyper}.
When computing topological features, we adopt graph batch processing to merge multiple batches of graph data into one large graph.
This approach solves the problem of constructed graph structures that need to be sparse due to fewer variables in the dataset.
Consequently, batch size becomes a hyperparameter that balances efficiency and accuracy. 
The calculation of topological features employs a multi-view approach, utilizing four different embedding functions for the Persistence Barcode.
Each function is applied three times, corresponding to the number of graph viewpoints.
\begin{table}[ht]
\caption{Default Hyperparameters Settings.}
\centering
\begin{tabular*}{\linewidth}{c|c|c|c}
\toprule
\textbf{Name} & \textbf{Value} & \textbf{Name} & \textbf{Value} \\
\midrule
Batch size & 32 & Top-K & 20 \\
Window size & 100 & Node embeddings & 128 \\
Window stride & 10 & Graph filtrations & 8 \\
Views of graph (k) & 12 &  Output layers & 2\\
\bottomrule
\end{tabular*}
\label{table: Hyper}
\end{table}

\section{Experimental Results}
In this section, we analyze the experimental results and present the visualization of graphs and their topologies.

\subsection{Results Analysis}
We compare traditional and state-of-the-art models from the past five years. 
The experiment results are shown in Table \ref{table:experiments}.
TopoGDN optimally balances accuracy and recall, achieving the highest F1-Score on all four datasets.
While some methods outperform TopoGDN in terms of precision or recall for specific datasets, the overall performance of TopoGDN is impressive.
TopoGDN outperforms the second-ranked model by about 2-3\% in the F1-Score on the SWaT and SMD datasets. 
This performance showcases its robustness and effectiveness in handling complex data scenarios.
The WADI dataset, characterized by a significant period and double to triple the number of multivariate features compared to other datasets, shows that all models' effects are less pronounced. 
ImDiffusion runs on the WADI dataset for over 8 hours without producing results due to the long runtime.
Nevertheless, TopoGDN outperforms all baseline models in accuracy and F1-Score, confirming that our model is suitable for large-scale datasets with some generalization ability.

\begin{table}[ht]
    \centering
    \caption{Model efficiency analysis for TopoGDN, TranAD, and ImDiffusion.}
    \begin{minipage}{\columnwidth}
    \centering
    \begin{tabular}{l|lll}
        \toprule
        \textbf{Model} & TopoGDN & TranAD & ImDiffusion \\
        \midrule
        \textbf{Total params (K)} & 592 & 619 & 4966 \\
        \textbf{Trainable params (K)} & 591 & 617 & 4823 \\
        \textbf{Params size (KB)} & 227 & 346 & 1806 \\
        \textbf{Flops (K)} & 460 & 672 & 4352 \\
        \textbf{Running Time (h)} & 1.2 & 2.3 & 7.8 \\
        \bottomrule
    \end{tabular}
    \label{tab:model_comparison}
    \end{minipage}
\end{table}
The efficiency comparative analysis shown in Table \ref{tab:model_comparison} focuses on two models, TranAD and ImDiffusion, which rank just behind TopoGDN in terms of their performance measured by the F1-Score. 
This comparison aims to evaluate the effectiveness and efficiency of the models in terms of model parameters and computational effort. 
We count the number of model parameters with a batch size of 1 and the number of trainable parameters, excluding static parameters such as position embeddings and graph structure information.
We also calculate the model's FLOPs (floating-point operations per second) per input unit and find a positive correlation with the inference time.
Moreover, we save the model parameters using a pre-training mode, count their sizes on the SMD dataset, and calculate the model's running time.
The efficiency test results demonstrate that TopoGDN is optimal regarding time and space complexity. 
The number of parameters for TopoGDN primarily arises from the TCN and GAT, which handle extraction timings and inter-feature dependencies, respectively.
For TranAD, the parameters are found mainly in the Transformer's multi-head mechanism and the feed-forward network within the Encoder-Decoder architecture. 
In ImDiffusion, the parameters primarily reside in the convolution operations of the noise predictor and the dimensionality involved in the transformation process. 
The experimental results indicate that TopoGDN has the optimal number of model parameters.
The computational workload of TopoGDN lies in the extraction and transformation of topological features and the backpropagation of the neural network. 
For TranAD, the main computational effort involves confronting the two losses during training. 
As a result, TranAD's time complexity is higher, and its inference time is 2-3 times that of TopoGDN. 
ImDiffusion's denoise process in the Diffusion model samples is too frequent, resulting in inference times more than 7-8 times longer than TopoGDN, making it impractical for large-scale datasets such as WADI.
\begin{table*}[ht]
    \caption{Anomaly Detection Performance in terms of precision(\%), recall(\%) and F1-score on four datasets. \textmd{The best results are bolded, and the runner-up results are underlined. The last column calculates the average F1-Score on the four datasets.}}
    \centering
    \large
    \begin{tabularx}{\textwidth}{@{}l|ccc|ccc|ccc|ccc|c@{}}
        \toprule
        \multirow{2}{*}{\textbf{Method}} & \multicolumn{3}{c|}{\textbf{MSL}} & \multicolumn{3}{c|}{\textbf{SWaT}} & \multicolumn{3}{c|}{\textbf{SMD}} & \multicolumn{3}{c|}{\textbf{WADI}} & \multirow{2}{*}{\shortstack{\textbf{Avg} \\ \textbf{F1}}} \\
        \cmidrule(r){2-4} \cmidrule(lr){5-7} \cmidrule(lr){8-10} \cmidrule(l){11-13}
        & \textbf{Prec} & \textbf{Rec} & \textbf{F1} & \textbf{Prec} & \textbf{Rec} & \textbf{F1} & \textbf{Prec} & \textbf{Rec} & \textbf{F1} & \textbf{Prec} & \textbf{Rec} & \textbf{F1} & \\ 
        \midrule
        \textbf{PCA}          & 12.62 & 56.12 & 20.61 & 24.92 & 21.53 & 23.10 & 27.18 & 34.57 & 30.50 & 39.18 & 5.93 & 9.86 & 21.02 \\
        \textbf{LSTM-VAE}     & 85.49 & 79.94 & 82.62 & 77.78 & 51.09 & 61.67 & 75.76 & 90.08 & 82.30 & 25.13 & 73.19 & 37.41 & 66.00 \\
        \textbf{MAD-GAN}      & 81.16 & 98.30 & 88.91 & 72.63 & 52.64 & 61.03 & \textbf{97.79} & 81.44 & 88.87 & 22.33 & \textbf{91.24} & 35.88 & 68.67 \\
        \textbf{OmniAnomaly}  & 89.02 & 86.37 & 87.67 & 52.57 & \textbf{95.46} & 67.80 & 83.68 & 86.82 & 85.22 & 31.58 & 65.41 & 42.60 & 70.82 \\
        \textbf{GDN}          & 78.75 & \textbf{99.81} & 88.04 & 84.86 & 58.17 & 69.02 & 69.36 & 71.14 & 70.30 & \textbf{59.80} & 30.77 & 40.61 & 66.99 \\
        \textbf{TranAD}       & \textbf{92.26} & 86.68 & \underline{89.38} & \underline{87.51} & 69.85 & \underline{77.69} & 84.62 & \textbf{94.71} & \underline{89.38} & 35.29 & \underline{82.96} & \underline{49.51} & \underline{76.49} \\
        \textbf{ImDiffusion}  & \underline{89.30} & 86.68 & 87.79 & 78.89 & \underline{74.62} & 76.70 & 85.20 & \underline{91.06} & 88.03 & - & - & - & - \\
        \textbf{TopoGDN (Ours)} & 82.37 & \underline{99.67} & \textbf{90.23} & \textbf{87.93} & 71.91 & \textbf{79.11} & \underline{97.57} & 83.39 & \textbf{89.92} & \underline{58.08} & 44.85 & \textbf{50.59} & \textbf{77.46} \\
        \bottomrule
    \end{tabularx}
    \label{table:experiments}
\end{table*}
\subsection{Ablation Study}
\begin{table*}[ht]
\caption{Ablation results of TopoGDN modules.}
\centering
\large
\begin{tabularx}{\linewidth}{@{}l|ccc|ccc|ccc|ccc|c@{}}
        \toprule
        \multirow{2}{*}{\textbf{Method}} & \multicolumn{3}{c|}{\textbf{MSL}} & \multicolumn{3}{c|}{\textbf{SWaT}} & \multicolumn{3}{c|}{\textbf{SMD}} & \multicolumn{3}{c|}{\textbf{WADI}} & \multirow{2}{*}{\shortstack{\textbf{Avg} \\ \textbf{F1}}} \\
        \cmidrule(r){2-4} \cmidrule(lr){5-7} \cmidrule(lr){8-10} \cmidrule(l){11-13}
                        & \textbf{Prec} & \textbf{Rec} & \textbf{F1} & \textbf{Prec} & \textbf{Rec} & \textbf{F1} & \textbf{Prec} & \textbf{Rec} & \textbf{F1} & \textbf{Prec} & \textbf{Rec} & \textbf{F1} \\ 
    \midrule
    GDN             & 78.75 & \underline{99.81} & 88.04 & 84.86 & 58.17 & 69.02 & 69.36 & 71.14 & 70.30 & 59.80 & 30.77 & 40.61 & 66.99 \\
    GDN + TCN        & 80.01 & 99.76 & 88.80 & 82.17 & \underline{62.43} & 70.95 & 93.06 & 67.34 & 78.10 & \underline{72.24} & 34.06 & 46.27 & 71.03\\
    GDN + MSTCN     & \underline{82.20} & \textbf{99.87} & \underline{90.21} & 83.67 & 61.38 & 70.81 & \underline{94.38} & 74.00 & 82.91 & \textbf{74.34} & 35.06 & \underline{47.62} & 72.89\\
    GDN + TA & 80.45 & 97.76 & 88.26 & \underline{89.00} & 60.51 & \underline{72.04} & 86.53 & \textbf{83.52} & \underline{84.95} & 57.24 & \underline{40.56} & 47.48 & \underline{73.18}\\
    TopoGDN (Ours)         & \textbf{82.37} & 99.67 & \textbf{90.23} & \textbf{92.01} & \textbf{65.04} & \textbf{76.21} & \textbf{97.57} & \underline{83.39} & \textbf{89.92} & 58.08 & \textbf{44.85} & \textbf{50.59} & \textbf{76.73}\\
    \bottomrule
\end{tabularx}
\label{table:performance}
\end{table*} 
Our model refers to the overall framework of the GDN, exhibiting significant improvement over the original model across all four datasets shown in Table \ref{table:performance}. 
Specifically, integrating all modules improves the average F1-Score on all datasets by 7\% to 19\%, respectively.
TopoGDN integrates a Multi-Scale Temporal Convolution (MSTCN) and a Topological Analysis module (TA) into the standard GDN framework. 
It also performs the Softmax normalization before the element-wise multiplication in the output layer.
We also compare the effects of single-scale Temporal Convolution (TCN) and Multi-Scale Temporal Convolution (MSTCN). 
Adding TCN and MSTCN modules improves the model's overall effectiveness, with an average F1-Score improvement of about 6\%, which can reach 12\% on the SMD dataset. 
This indicates that the SMD dataset exhibits richer temporal features.
MSTCN has a slight advantage in experiment results compared to single-scale TCN.
On the SWaT, WADI, and SMD datasets, the Topological Analysis module markedly improves the model's performance, with enhancements ranging from 4\% to 17\% in the F1-Score. 
This indicates that the TA module can substantially boost the feature extraction capabilities of GNNs. 
This module has been proven to integrate into various neural network architectures such as GCN, GAT, and Graph Isomorphism Network (GIN), consistently achieving significant enhancements \cite{horn2021topological, moor2020topological}.
These modules are linked via residual connections and are designed to be plug-and-play, facilitating their integration into other GNNs and temporal networks to enhance spatiotemporal feature extraction.

\subsection{Hyperparameter Sensitivity}
\begin{figure*}[ht]
  \centering
  \begin{subfigure}[t]{.24\linewidth} 
    \centering
    \includegraphics[width=1.1\linewidth, height=0.5\linewidth]{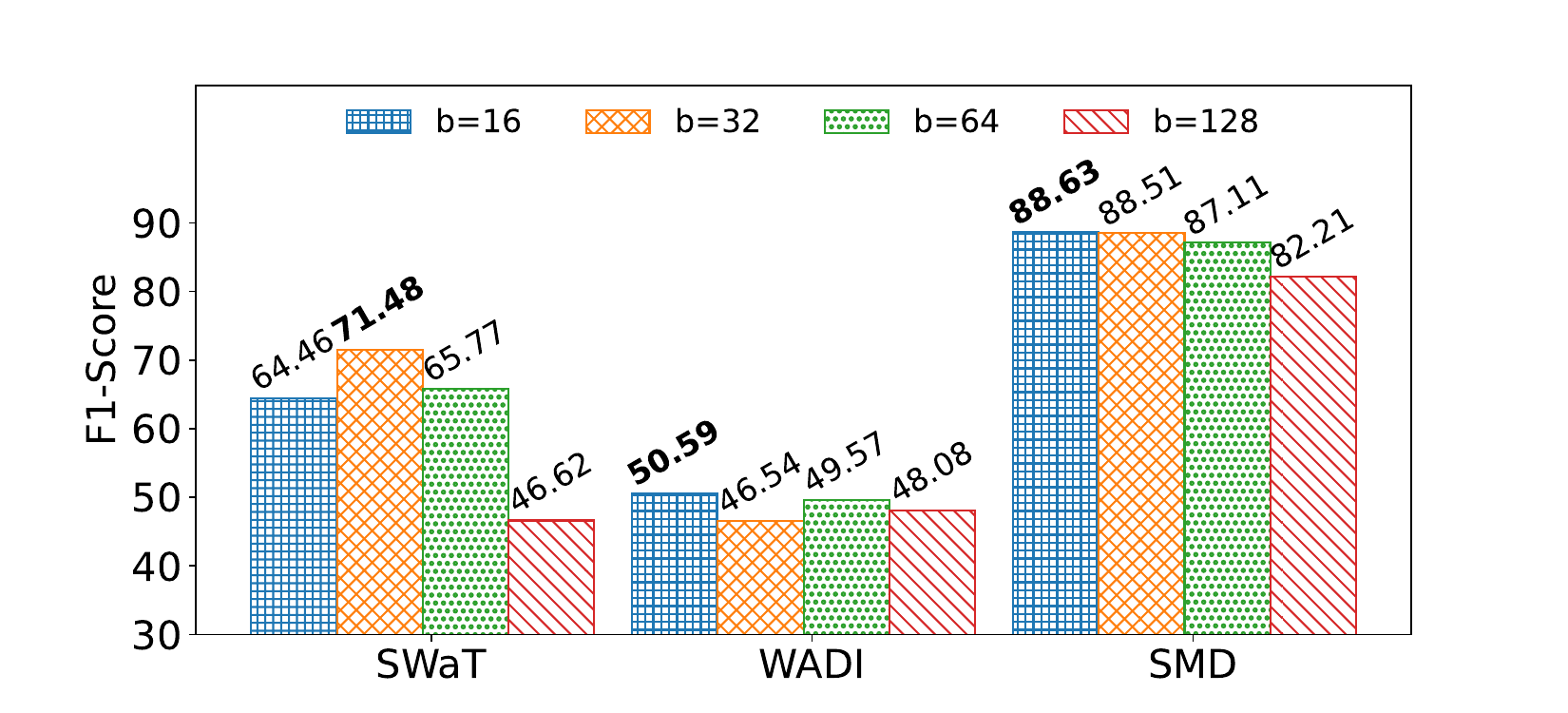}
    \caption{Batch Size (b).}
    \label{bs}
  \end{subfigure}
  \begin{subfigure}[t]{.24\linewidth}
    \centering
    \includegraphics[width=1.1\linewidth, height=0.5\linewidth]{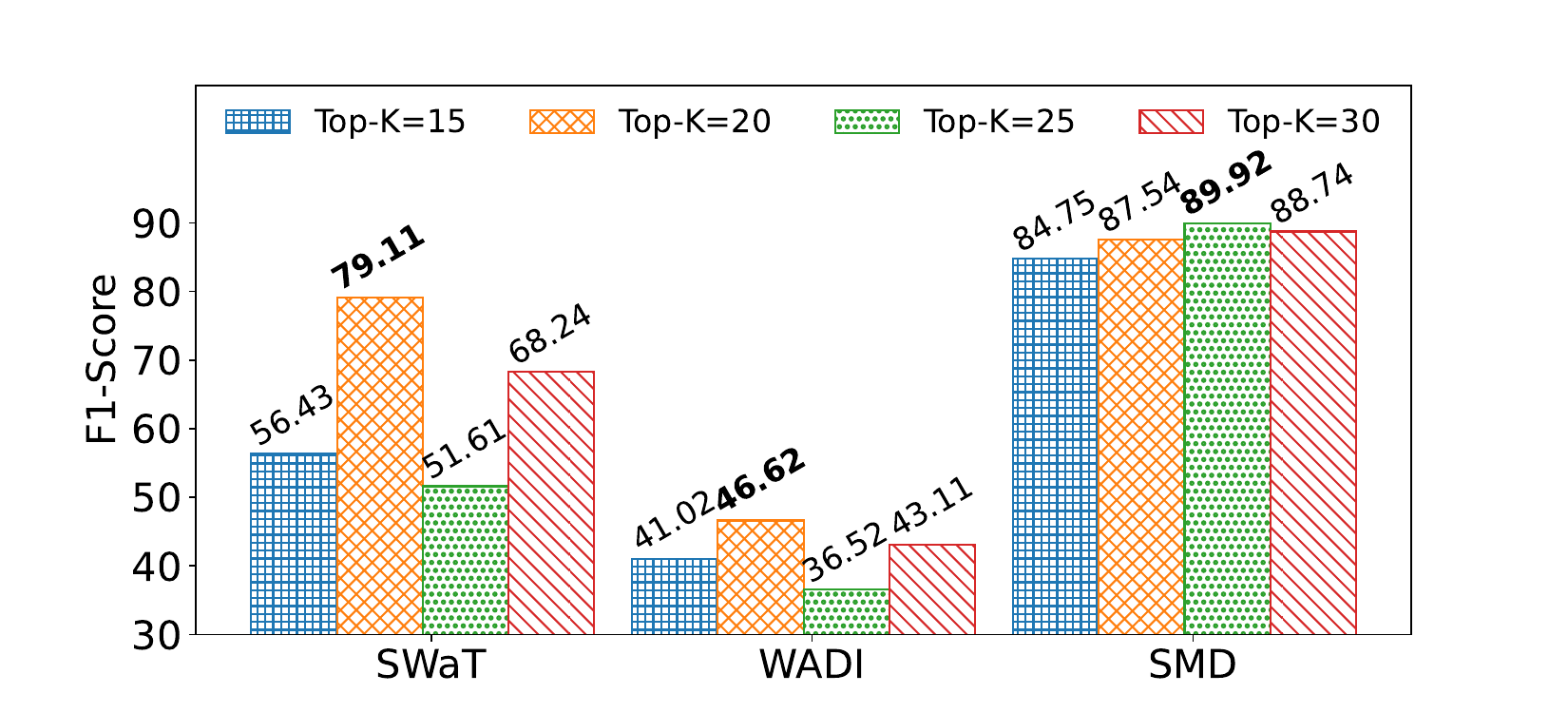} 
    \caption{Top-K.}
    \label{topk}
  \end{subfigure}
  \begin{subfigure}[t]{.24\linewidth}
    \centering
    \includegraphics[width=1.1\linewidth, height=0.5\linewidth]{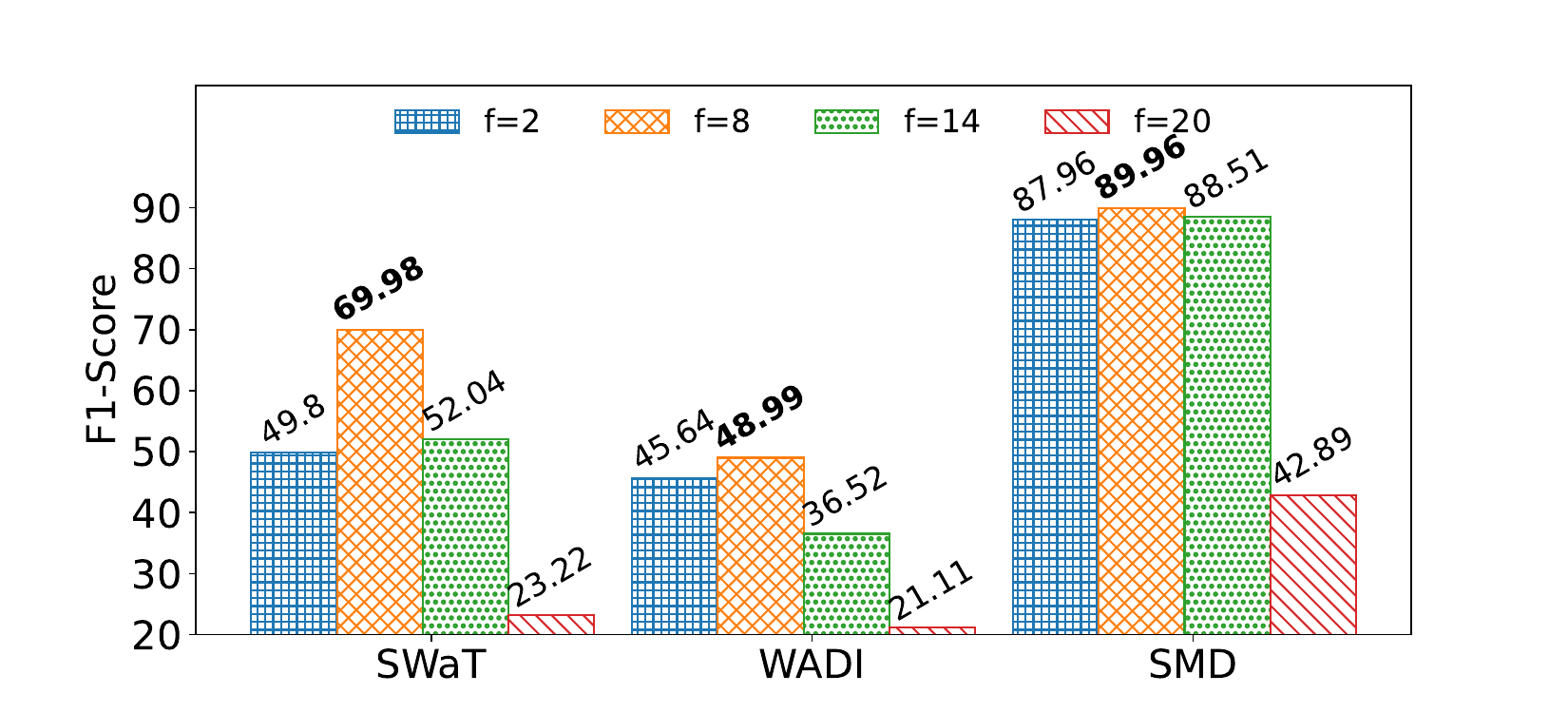}
    \caption{Graph filtrations (f).}
    \label{f}
  \end{subfigure}
  \begin{subfigure}[t]{.24\linewidth}
    \centering
    \includegraphics[width=1.1\linewidth, height=0.5\linewidth]{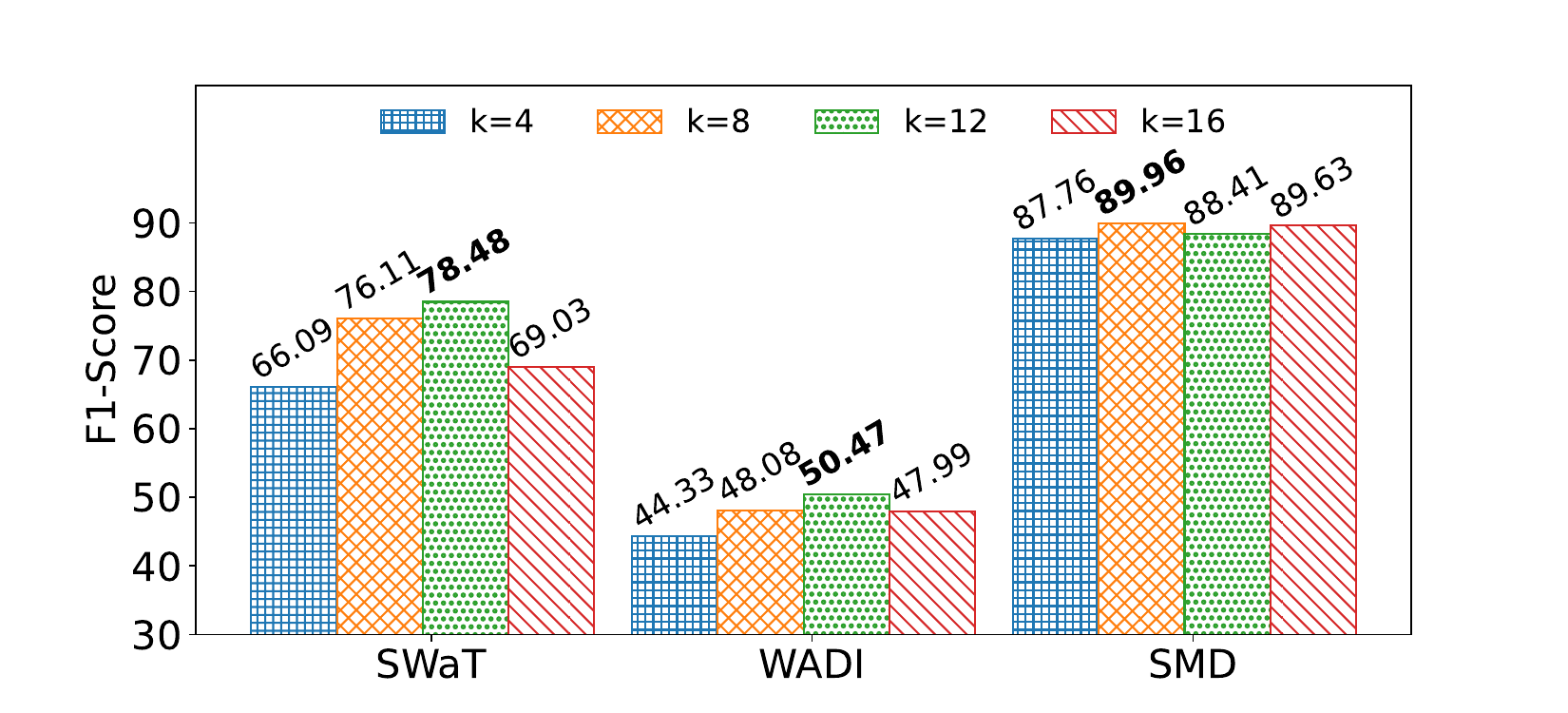}
    \caption{Views of graphs (k).}
    \label{k}
  \end{subfigure}
  
  \begin{subfigure}[t]{.24\linewidth} 
    \centering
    \includegraphics[width=1.1\linewidth, height=0.5\linewidth]{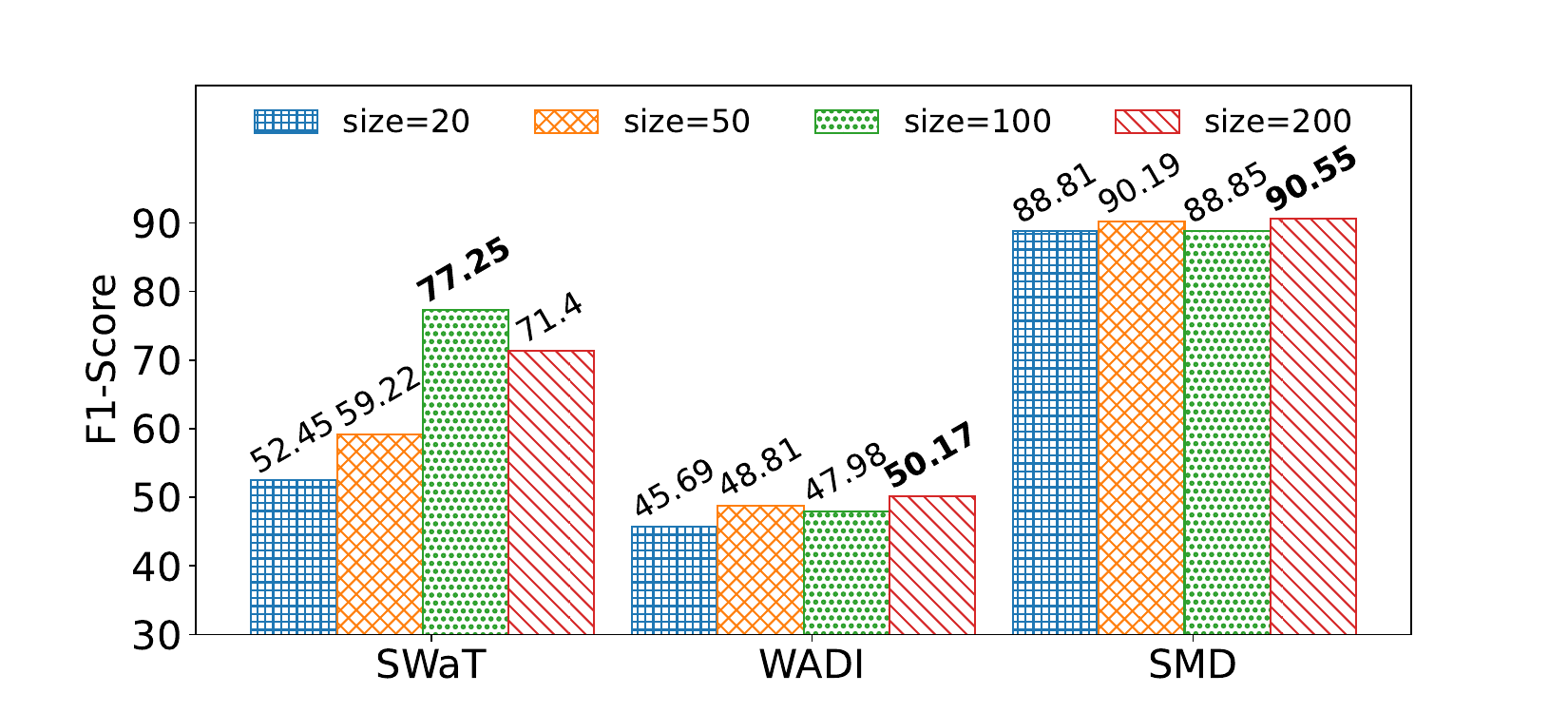} 
    \caption{Window size (size).}
    \label{window size}
  \end{subfigure}
  \begin{subfigure}[t]{.24\linewidth} 
    \centering
    \includegraphics[width=1.1\linewidth, height=0.5\linewidth]{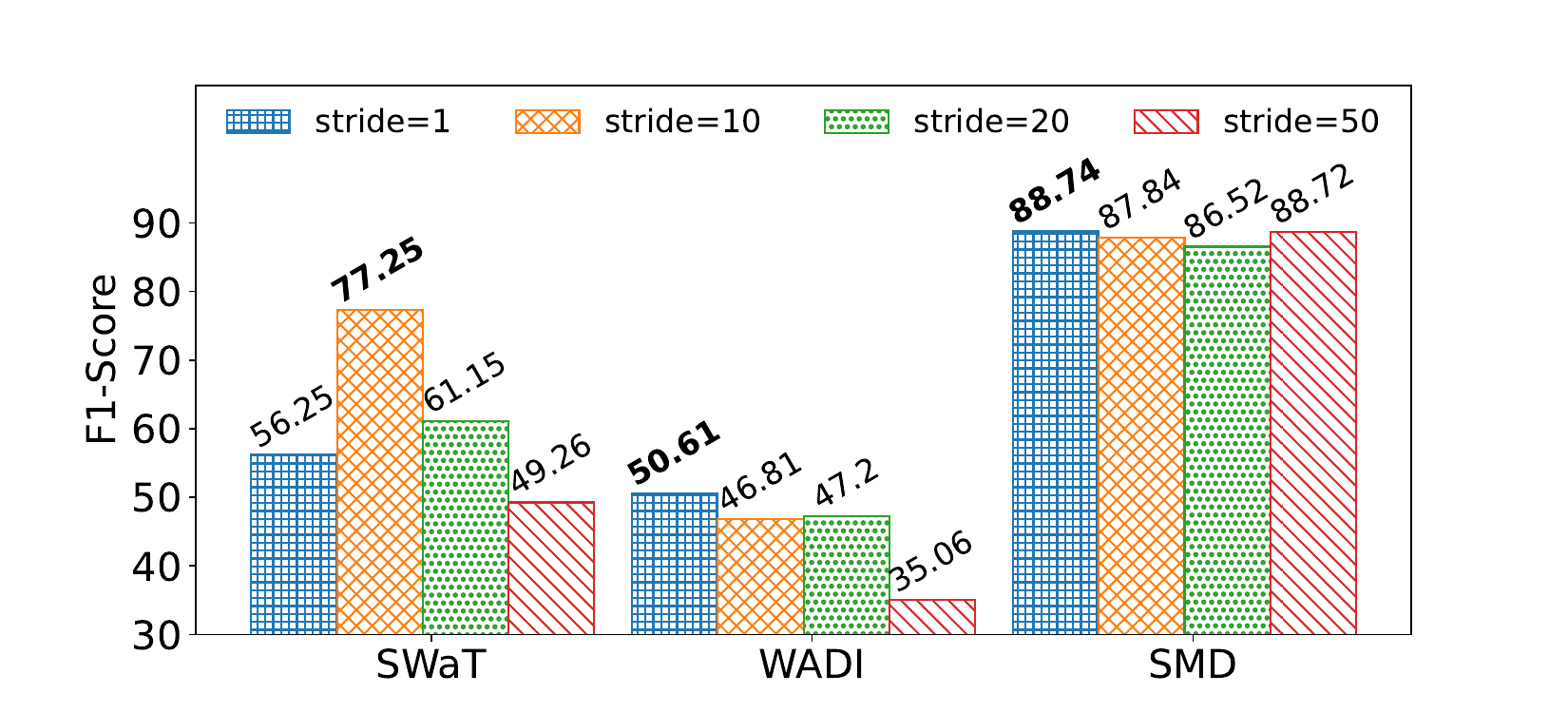} 
    \caption{Window stride (stride).}
    \label{window stride}
  \end{subfigure}
  \begin{subfigure}[t]{.24\linewidth} 
    \centering
    \includegraphics[width=1.1\linewidth, height=0.5\linewidth]{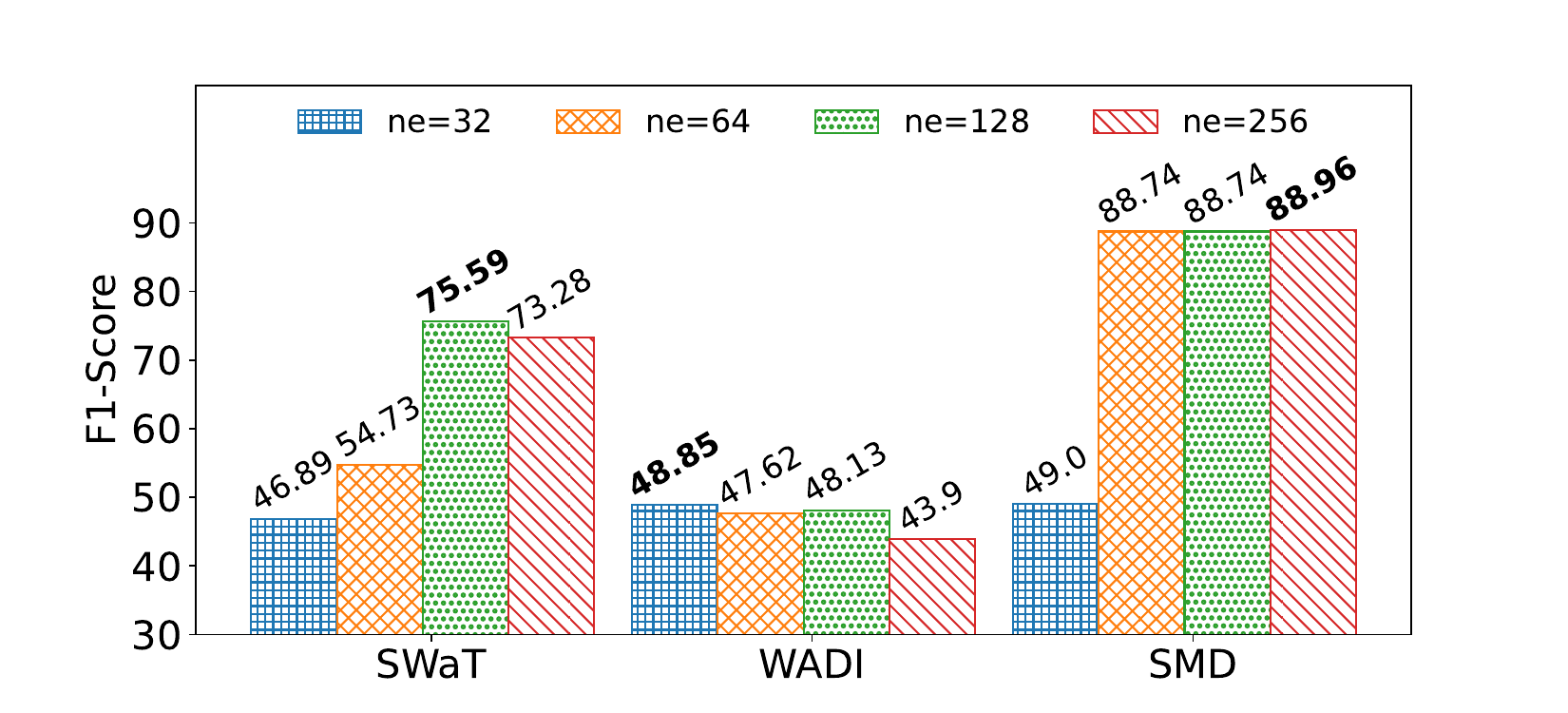} 
    \caption{Node embeddings (ne).}
    \label{ne}
  \end{subfigure}
  \begin{subfigure}[t]{.24\linewidth} 
    \centering
    \includegraphics[width=1.1\linewidth, height=0.5\linewidth]{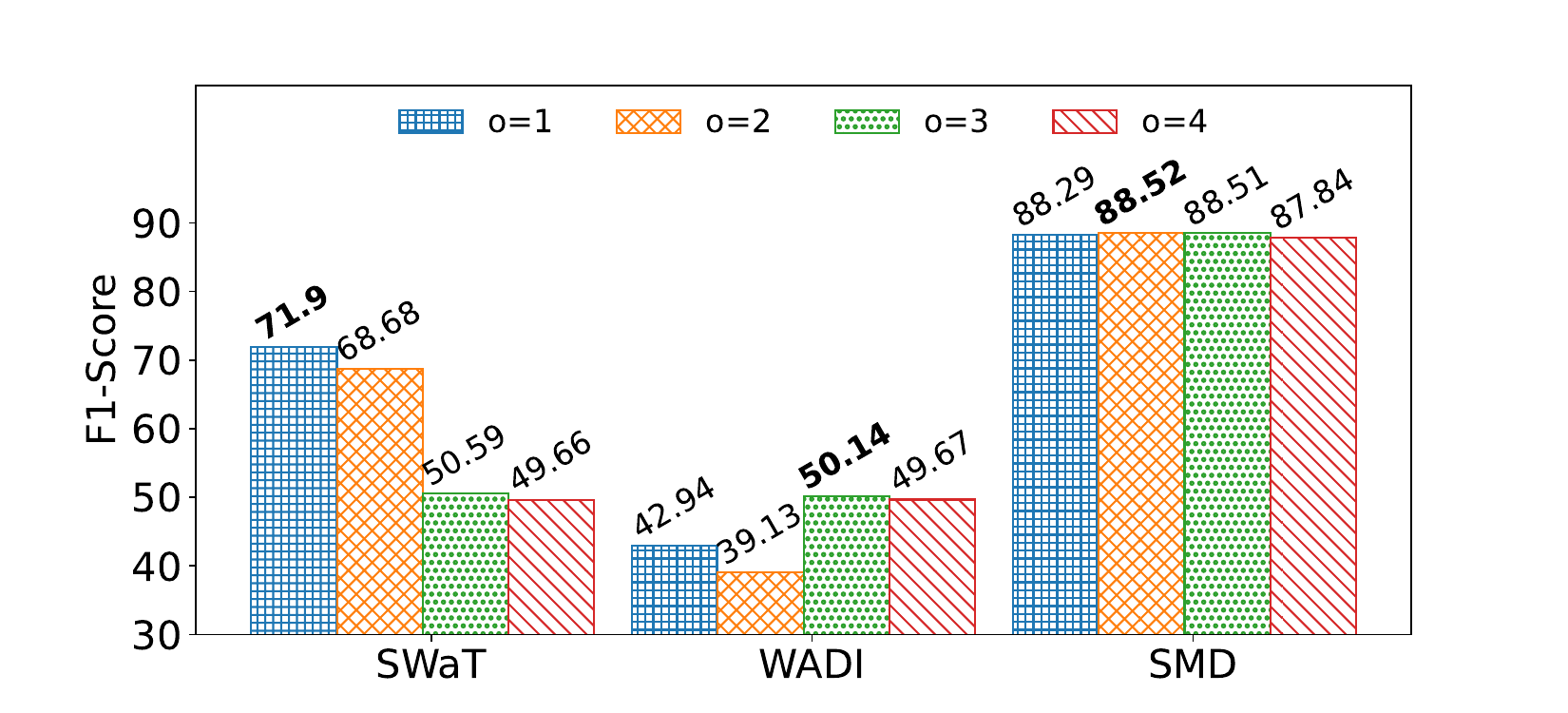} 
    \caption{Output layers (o).}
    \label{o}
  \end{subfigure}
  \caption{Effect of different hyperparameters on model performance.}
  \label{fig:HyperParameters}
\end{figure*}

We conduct experiments to evaluate the effects of different hyperparameters, as illustrated in Figure \ref{fig:HyperParameters}.
The batch size, depicted in Figure \ref{bs}, is crucial since we use one large fused graph instead of multiple small graphs for batch processing.
For larger datasets, a smaller batch size, typically around 16, ensures a moderately dense graph structure within the batch. 
If the batch size is too large, the topological features after graph filtering become less distinct. 
The Top-K parameter, shown in Figure \ref{topk}, representing the number of edges per node, also impacts topological feature extraction. 
Connecting too many edges makes it difficult to extract distinct topological features and values of 15 or 20, which are commonly used.
Excessive graph filtering operations can destroy structural features, as indicated in Figure \ref{f}, where the detection effectiveness across all three datasets declines dramatically when graph filtration is performed 20 times.
The window size and stride are dependent on the dataset size. 
A larger dataset exhibits the richer temporal feature. 
As shown in Figure \ref{window size} and \ref{window stride}, the model is more effective on the WADI dataset when the window size is set to a maximum value of 200 and the stride is minimized to 1.
The choice of node embedding and the number of output layers also depends on the dataset. 
Generally, larger values increase the likelihood of overfitting.
As demonstrated in Figure \ref{ne} and \ref{o}, overfitting is observed to varying degrees across all three datasets.

\subsection{Topological Feature Visualization}
\begin{figure}[htbp]
  \centering
  \begin{subfigure}[b]{0.48\textwidth}
    \includegraphics[width=\textwidth]{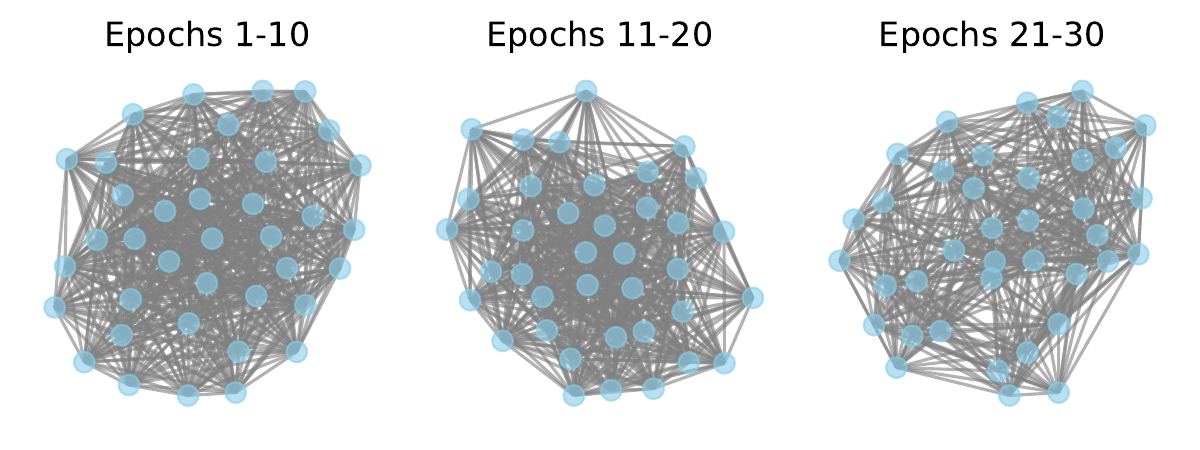}
    \caption{Graph Structure for Top-K=15.}
    \label{fig:sub1}
  \end{subfigure}
  \hfill
  \begin{subfigure}[b]{0.48\textwidth}
    \includegraphics[width=\textwidth]{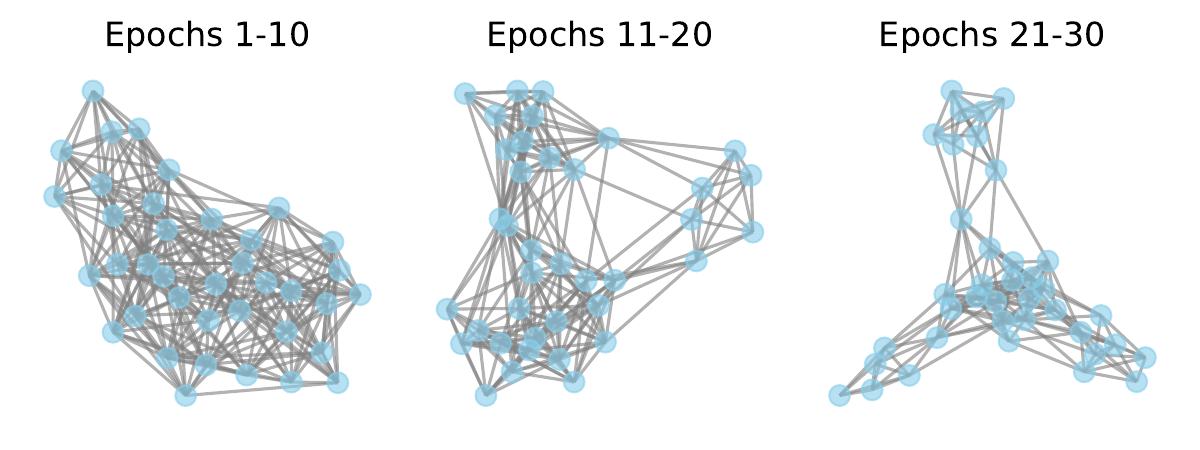}
    \caption{Graph Structure for Top-K=5.}
    \label{fig:sub2}
  \end{subfigure}
  \caption{Visualization of Graph Structure.}
  \label{fig:graph structure}
\end{figure}


The topological analysis module extracts higher-order features from the learned graph structure information. 
Figure \ref{fig:graph structure} illustrates the graph structure obtained from training on the SMD dataset with different Top-K values. 
The results show that when Top-K = 5, the graph tends to be sparser. 
The nodes of the graph form clusters of multiple types along the training, corresponding to the two data types in the SMD dataset: binary data with values 0 and 1, and flow data, which shows different data distributions in [0,1] after normalization.
\begin{figure}[!ht]
    \centering
    \includegraphics[width=0.48\textwidth]{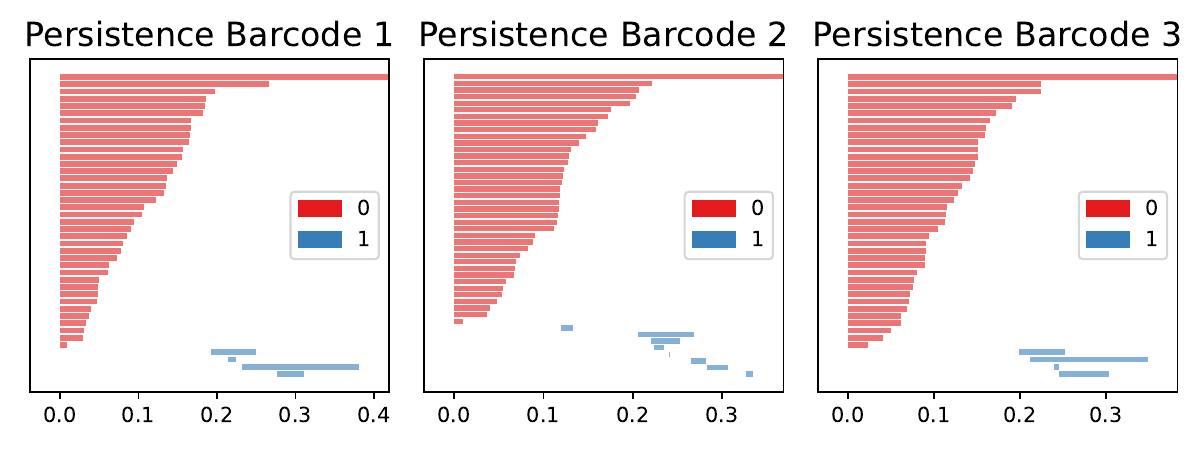}
    \caption{Visualization of Persistence Barcode.}
    \label{fig:barcode}
\end{figure}

Figure \ref{fig:barcode} displays a visualization of the Persistence Barcode computed for the graph structures obtained from different training processes at Top-K = 5. 
Each horizontal line (a bar) represents a topological feature (e.g., a connectivity component or a ring) across a specific range of scales. 
The beginning of the bar indicates the scale at which the feature first appears, and the end suggests the scale at which the feature disappears. 
The red bar represents a 0-dimensional homology group, typically corresponding to a connected component. 
The length of the bar indicates the duration from when the connected component is formed until it eventually merges with other elements. 
Blue bars represent 1-dimensional homology groups corresponding to rings or holes in the data. 
The start and end points of the blue bars denote when the ring structure forms and disappears. 
Many short-lived red bars in the first of these plots suggest that many connected components merge rapidly. 
A few long bars indicate that some connected components remain independent on larger scales. 
There are fewer blue bars, indicating the presence of a small number of persistent ring structures. 
The second graph displays more blue bars than the previous one, suggesting more persistent ring structures, and the graph's structural information becomes more evident as training progresses. 
The red bars have a similar distribution but are clustered on a shorter scale, indicating quicker merging of the connected components. 
The third graph shows almost no blue bars, while the red bars are numerous and primarily short-lived, suggesting that connected components form and merge quickly in the data, with little to no persistent ring structure.
\section{Conclusion}
We propose a prediction-based multivariate time series anomaly detection method called TopoGDN.
This method extracts temporal features using multi-scale temporal convolution with kernels of different sizes. 
Concurrently, the model captures cross-feature dependencies through the graph structure learning and GAT. 
Furthermore, the method incorporates a plug-and-play topological analysis module to integrate higher-order structural information at multiple scales. 
This process significantly enhances GAT's feature extraction capabilities. 
Experimental results demonstrate that our method outperforms other baseline models on four datasets.

\begin{acks}
This work is supported by the National Key R\&D Program of China through grant 2021YFB1714800, NSFC through grant 62322202, Beijing Natural Science Foundation through grant 4222030, Shijiazhuang Science and Technology Plan Project through grant 231130459A, and Guangdong Basic and Applied Basic Research Foundation through grant 2023B1515120020.
\end{acks}

\bibliographystyle{ACM-Reference-Format}
\balance
\bibliography{references}

\end{document}